\documentclass{article}

\usepackage[final]{nips_2017}

\usepackage[utf8]{inputenc} 
\usepackage[T1]{fontenc}    
\usepackage{hyperref}       
\usepackage{url}            
\usepackage{booktabs}       
\usepackage{amsfonts}       
\usepackage{nicefrac}       
\usepackage{microtype}      
\usepackage{graphicx}
\usepackage{subfigure}

\usepackage{algorithm}
\usepackage[noend]{algorithmic}

\usepackage{bm}

\usepackage{cite,amsfonts,amssymb,amsmath,exscale,bbm}
\makeatletter
\newcommand\footnoteref[1]{\protected@xdef\@thefnmark{\ref{#1}}\@footnotemark}
\makeatother

\hypersetup{ %
    pdftitle={Adversarial Learning of Adversarially Robust Representation},
    pdfkeywords={},
    pdfborder=0 0 0,
  pdfpagemode=UseNone,
    colorlinks=true,
    linkcolor=blue, 
    citecolor=blue, 
    filecolor=blue, 
    urlcolor=blue, 
    pdfview=FitH,
    pdfauthor={Anonymous},
}

\newcommand{\beq}{\begin{equation}}
\newcommand{\eeq}{\end{equation}}
\newcommand{\be}{\begin{equation}}
\newcommand{\ee}{\end{equation}}
\newcommand{\beqa}{\begin{eqnarray}}
\newcommand{\eeqa}{\end{eqnarray}}
\newcommand{\bean}{\begin{eqnarray*}}
\newcommand{\eean}{\end{eqnarray*}}

\renewcommand{\omega}{w}

\newcommand{\R}{\mathbb{R}}

\def\maps{\colon}

\def\to{\rightarrow}

\DeclareMathOperator*{\argmax}{arg\,max}

\newcommand*\samethanks[1][\value{footnote}]{\footnotemark[#1]}

\title{A3T: Adversarially Augmented \\ Adversarial Training}

\author{
  Akram Erraqabi\thanks{equal contribution}  ~~~ Aristide Baratin\samethanks ~~~ Yoshua Bengio ~~~Simon Lacoste-Julien
    \\
  Montreal Institute of Learning Algorithms\\
  Université de Montréal\\
  Montreal, Canada \\
}
\begin{document}

\maketitle

\begin{abstract}
Recent research showed that deep neural networks are highly sensitive to so-called adversarial perturbations, which are tiny perturbations of the input data purposely designed to fool a machine learning classifier. 
In this work, we investigate a procedure to improve adversarial robustness of deep neural networks through enforcing representation invariance. The idea is to train the classifier jointly with a discriminator attached to one of its hidden layer and trained to filter the adversarial noise. We perform preliminary experiments to test the viability of the approach and to compare it to other standard adversarial training methods. 

\end{abstract}

\section{Introduction}

The recent impressive advances of machine learning and in particular deep learning methods in solving several challenging tasks, came along with a vulnerability that questions the security of several applications. In fact, as shown in ~\citet{Szegedy2013}, small and visually imperceptible perturbations of images, called \emph{adversarial examples}, can lead to highly confident misclassification. Moreover, these 
adversarial examples can generalize across different networks ~\citep{papernot2016practical, Universal_adv}, which gives the \emph{adversary} the ability to fool artificial systems without any prior knowledge on their architecture and makes models deployed for real-world applications an easy target of adversarial attacks.


A growing body of work is now devoted to designing adversarial attacks, quantifying the robustness of classifiers  and building effective defense mechanisms ~\citep{Fawzi2015, Kurakin2016, papernot2016distillation, DeepFool, Ensemble_adv}. A standard approach known as \emph{adversarial training}, suggested in \citet{Goodfellow2015_adv}, consists in augmenting the training set with its adversarial version. More recent methods \citep{mixup} try to encourage linear behavior between training examples, thus building some robustness to adversarial examples.

In this paper, we suggest a new approach to building some resistance to adversarial attacks by training a discriminator to distinguish latent representations of real samples from the ones of adversarial examples. The discriminator gets intermediate features of a classifier as input. At the same time, the classifier is trained not only to correctly classify the training data but also to fool the discriminator when fed an adversarial example. This enforces invariance of the hidden representation with respect to the nature of the input i.e whether it is a real sample or an adversarial examples. The classifier ends up filtering out the adversarial \emph{noise} at the level of the hidden representation, leading the adversarial example to be classified as its corresponding real sample.    

We perform experiments on the MNIST datasets to validate the proof of concept. Our results suggest that our procedure compares well to standard adversarial training, and that combining the two methods may further improve robustness.

\section{Related work}

The idea of augmenting a network with a discriminator to enforce hidden representation invariance is at the heart of the work \citet{Ganin2016domainadaptation} on domain adaptation, where the network learns features that adapt to different domains for the same task.
This idea has been exploited in several recent works.
\citet{metzen2017detecting} uses this kind of architecture in the context of {\it detection}: the discriminator, trained separately from the classifier, is merely  used as detector of adversarial attacks. 
By contrast,  the role of the discriminator in our approach is to teach the classifier to build robust features, in a procedure similar in spirit to the GAN framework ~\citep{Goodfellow2014GAN}.

Fader nets \citep{lample2017fader} use an adversarial loss for an encoder to extract the invariant patterns that are orthogonal to some image attributes. This helped disentangling these attributes factors  and allowed controlling them in a generative model setting.

\citet{nuisances} takes the view from information theory  to a regularizing term added to the classifier loss, which filters out the information of a given class of  `nuisances', i.e variations of the inputs that are  irrelevant to the task. This regularizer takes the form of mutual information, which can be estimated by means of a discriminator. Our work can be viewed as a generalization of this proposal to the case of adversarial perturbations.

\section{Approach}

\subsection{Adversarial perturbations} 

Consider a classifier $C_\theta \maps \mathcal{X} \to \mathcal{Y}$  
trained by minimizing  the average of a given loss function $\mathcal{L}_\theta(x, y)$ over labeled examples $(x, y)$ in a training dataset. 
An \emph{adversary} willing to fool the classifier may design input perturbations $\bar{x} = x+\delta(x)$ so as to maximize the loss: 
\beq \label{adv} 
\delta(x) = \argmax_{\delta \in \mathcal{C}} \mathcal{L}_\theta(x+\delta, y) 
\eeq
where $\mathcal{C}$ encodes a set of constraints controlling the amplitude of the perturbations -- e.g small enough to remain quasi-imperceptible to a human observer. As shown in ~\citet{Szegedy2013}, deep neural networks are highly unstable under such adversarial perturbations.

The simplest methods to generating perturbations defined in (\ref{adv}) are first order approaches,  which linearly expand the loss 
$\mathcal{L}_\theta(x+\delta, y)  \simeq \mathcal{L}_\theta(x, y) + \langle \delta, \nabla_x \mathcal{L}_\theta(x, y)\rangle$. If the constraints $\mathcal{C}$ take the form $\|\delta\|_{\infty} \leq \epsilon$ for some $\epsilon >0$, 
Equ. (\ref{adv}) yields the so-called  fast gradient method for generating adversarial examples ~\citep{Goodfellow2015_adv}: 
\beq \label{sign_grad}
x^{adv}  = x + \epsilon \mbox{sign} \left(\nabla_x \mathcal{L}_ \theta(x, y)\right)
\eeq
Other generation methods have been proposed, which include ~\citet{Kurakin2016, DeepFool}. In our preliminary experiments, we use the procedure (\ref{sign_grad}) to generate adversarial examples {\it on-the-fly} during training.

\subsection{Augmented network}
\begin{figure}[h]
\centering 
\includegraphics[width=9cm]{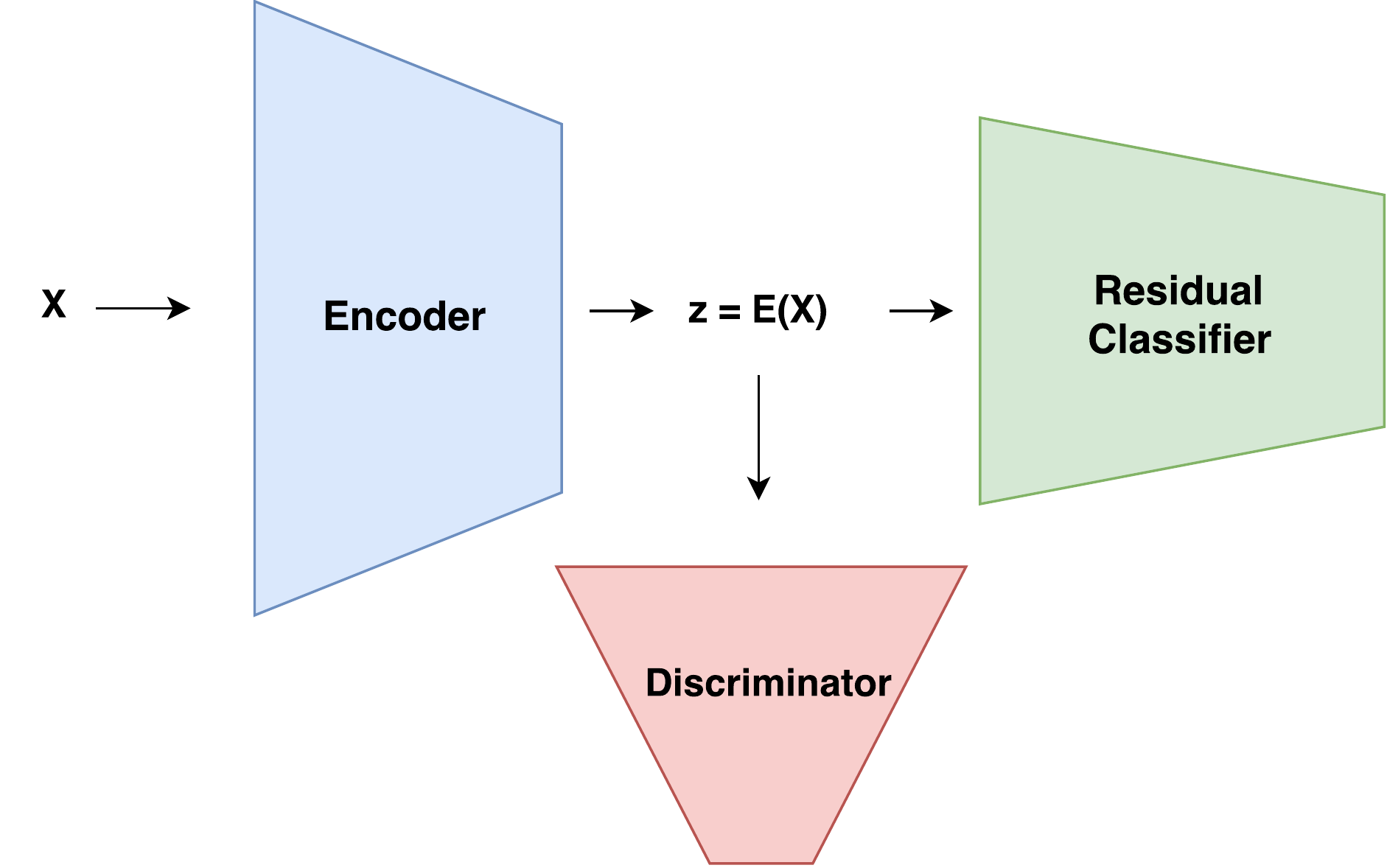}
\caption{Model architecture}
\label{fig:model_}
\end{figure}

Assuming the classifier has $L$ layers, we choose a layer $\ell$, which  effectively splits the network into an encoder part $E$ (layers $1, \cdots \ell$) and a residual classification part $R$ (layers $\ell, \ell+1, \cdots L$). We supplement the network with a discriminator $D$ branched off from $\ell$.  If  $\ell$ has $k$ units, $D \maps \R^k \to [0,1]$  takes as input the feature at $\ell$ and produces an output interpreted as the probability of the feature coming from a real (as opposed to adversarial) input. 
Given an input image $x \in \mathbb{R}^d$ we thus have the following notation: 
\begin{itemize} 
\item $z = E(x)$ is the feature vector at layer $\ell$
\item $R(z) = R(E(x)) = C_\theta(x)$ is the output of the original classifier  
\item $D(z)$ is the output of the discriminator, and the probability that $z$ is the representation of a sample of the real data. 
\end{itemize} 
The architecture is shown in Fig \ref{fig:model_}. 

\subsection{Adversarially Augmented Training} 

At each iteration, we forward-pass a mini-batch of real data $(x, y)$, then generate and forward pass the corresponding mini-batch of adversarial examples $(x^{adv}, y)$. 
Classifier and discriminator are trained simultaneously. 
The discriminator is trained as a binary classifier where $z$ has tag $t=1$ 
when  $z = E(x)$ and $t=0$ when $z=E(x^{adv})$. The classifier is trained to classify each real input correctly (which involves both encoder and residual) and to fool the discriminator (which involves only the encoder). Training the classifier on the discriminator's response should enforce an invariance across real samples and their adversarial versions at the level of the latent representation $z$.
This can be combined with standard adversarial training for the correct classification of each adversarial input.

Let $\theta_{enc}, \theta_{res}$ and $\theta_{disc}$ the parameters of the encoder, the residual classifier and the discriminator respectively, and $\theta=(\theta_{enc}, \theta_{res})$.  We use the following losses:

{\bf Classification Loss}. For each labeled sample $(x, y)$, the classifier outputs class probabilities $P_\theta(\cdot |x)$.  We use the cross-entropy loss for the classification task.  
This loss can be extended to include classical adversarial training \citep{Goodfellow2015_adv}:  
$$ \mathcal{L}_{\theta}(x,x^{adv}, y) = - \alpha \log P_\theta(y|x) - (1-\alpha)\log P_\theta(y|x^{adv}),$$
where $\alpha \in [0, 1]$ is a convex mixing parameter (we used $\alpha = 0.5$).

{\bf Discriminator Loss}. For each feature $z$ with $t$, the discriminator outputs the probability $D(z)=P_{\theta_{disc}}(t=1|z)$.  We use binary cross-entropy to train the discriminator to distinguish the real features from the adversarial ones:
$$
\mathcal{L}_{\theta_{disc}}(z, t) = -  \log P_{\theta_{disc}}(t|z)  
$$

{\bf Encoder Adversarial Loss}. We use the cross-entropy loss to train (the encoder part of) the classifier on the discriminator response, so as to make the adversarial representation be classified as real by the discriminator: 
$$ 
\mathcal{L}^{adv}_{\theta_{enc}}(x^{adv}) = -\beta \log D(E(x^{adv})),
$$
where $\beta > 0$ is a factor controlling the importance of the adversarial loss i.e how much we enforce invariance (we used $\beta = 1$).


\section{Experiments -- Proof of concept}
{\bf Setup.}
The reported experiments are based on a 3-layer feed-forward network using leaky ReLU as activations. The number of hidden units of its hidden layers are respectively 512, 256, 128. The discriminator is fed the output of the second hidden layer. The adversarial examples were generated using $\epsilon = 0.25$.

The discriminator is a simple shallow network with 128 hidden units, a ReLU activation and a 0.5-dropout.

{\bf Training.}
At each iteration, we update the discriminator, and the classifier with adversarially augmented loss. One can try multiple updates of the discriminator for one update of the classifier as suggested in \citet{Goodfellow2014GAN}, but we noticed that a single update gave already the expected training behavior.

\begin{figure}[h]
\centering
\includegraphics[width=13cm]{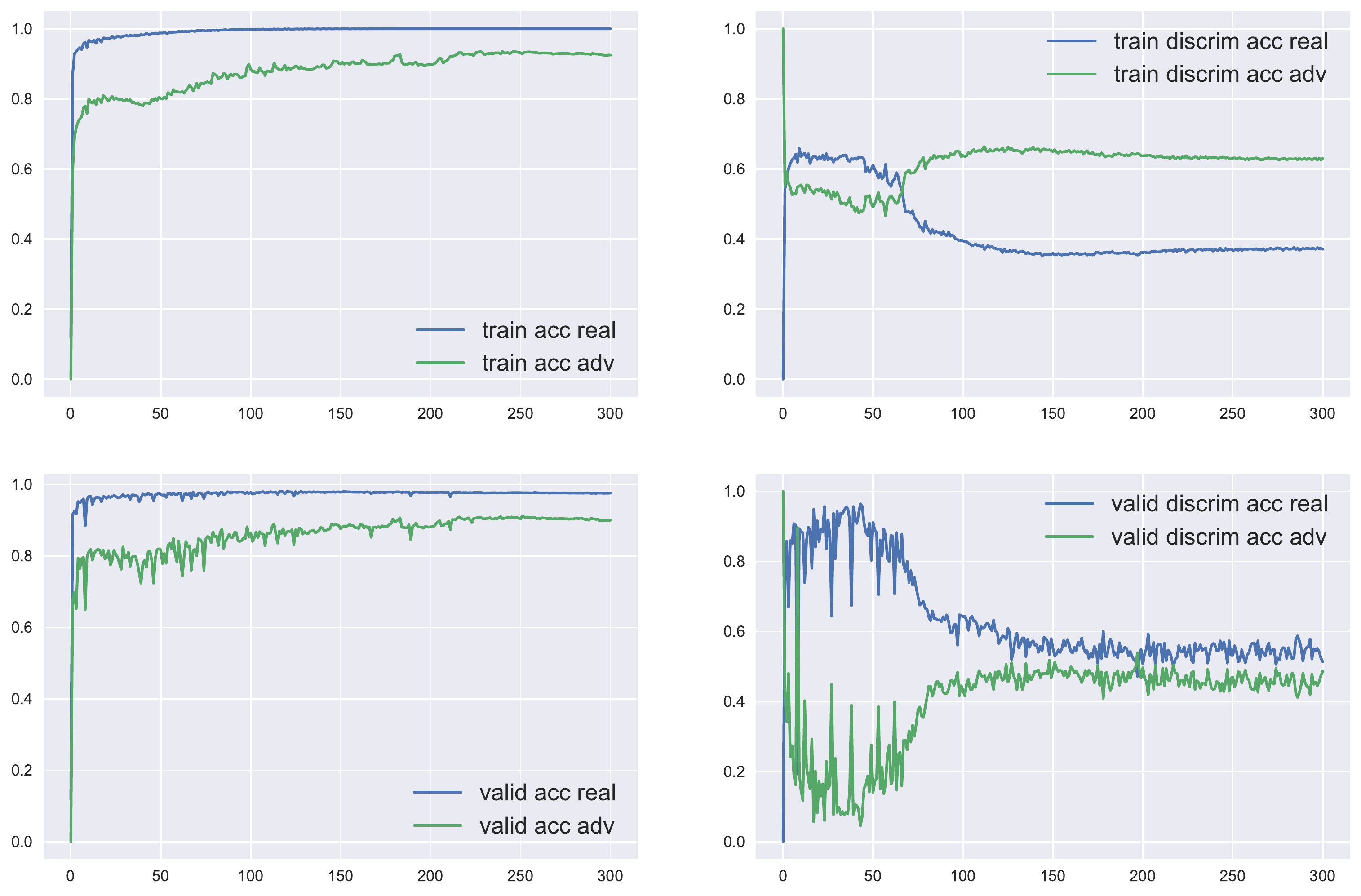}
\caption{Mnist Adversarially Augmented Training}
\label{fig:model_curves}
\end{figure}

Figure \ref{fig:model_curves} gives the classification accuracies reached by the classifier and the discriminator for real and adversarial examples in the training and validation dataset, as a function of the training epochs.  

Table \ref{tab:mnist_test} gives the test performance of the same model trained with 4 different approaches: simple training of the classifier,  standard adversarial training (SAT), adversarially augmented training (A2T) and finally the adversarially augmented adversarial training  (A3T). 
\begin{table}[h]
\centering
\begin{tabular}{ r || c | c }
   					& Accuracy on real & Accuracy on adversarial \\ \hline
  					Simple training & 98.18\% & 25.37\% \\
  					AT ($\alpha=\frac12, \beta=0$)  & 98.89\% & 94.45\% \\

       				A2T ($\alpha=\beta=1$) \hspace{3mm} & 97.77\% & 89.74\%\\
  	   A3T ($\alpha=\frac12, \beta=1$) & 98.72\% &  {\bf 96.10\%} \\
  					\hline  
					\end{tabular}
                    \vspace{5mm}
					\caption{Test accuracies for MNIST on real and adversarial samples ($\epsilon=0.1$, 10 runs)}
\label{tab:mnist_test}
\end{table}

\vspace*{-0.7cm}
\section{Discussion}


We proposed a new  approach to enforce invariance of a classifier's features with respect to adversarial perturbations. We proved the efficacy of the method with some proof of concept experiments.
More work needs to be done to test the sensitivity of the model to the hyperparameters and the adversarial generation method.  A crucial point will also be  to test its scalability to larger datasets and architectures.  

The existence of adversarial examples  suggests that neural networks, despite their impressive generalization abilities in perceptual tasks, do not learn the true underlying concepts defining the correct labels. One could hope that a procedure seeking feature invariance would push the network to not only ignore adversarial perturbations but also to focus more on relevant patterns in real data. We conjecture that our approach is one step in that direction. 



\bibliographystyle{apalike}
\bibliography{references}

\end{document}